\documentclass{article}
\pdfoutput=1

\PassOptionsToPackage{numbers}{natbib}
\usepackage[preprint]{neurips_2023}

\usepackage[utf8]{inputenc} 
\usepackage[T1]{fontenc}    
\usepackage{hyperref}       
\usepackage{url}            
\usepackage{booktabs}       
\usepackage{amsfonts}       
\usepackage{nicefrac}       
\usepackage{microtype}      
\usepackage{xcolor}         

\usepackage[export]{adjustbox}
\usepackage{amsmath}
\usepackage{amsthm}
\usepackage{bm}
\usepackage{mathtools}
\usepackage{multirow}
\usepackage{nicefrac}       
\usepackage{pifont}
\usepackage{pgfplots}
\usepackage{tikz}
\usepackage{subcaption}
\pgfplotsset{
    compat=1.17
}

\newcommand{\Bzero}{\bm{0}}

\newcommand{\Ba}{\bm{a}}

\newcommand{\Bh}{\bm{h}}

\newcommand{\Bp}{\bm{p}}

\newcommand{\Bv}{\bm{v}}
\newcommand{\Bx}{\bm{x}}

\newcommand{\By}{\bm{y}}

\newcommand{\BA}{\bm{A}}

\newcommand{\BD}{\bm{D}}
\newcommand{\BE}{\bm{E}}

\newcommand{\BI}{\bm{I}}
\newcommand{\BK}{\bm{K}}

\newcommand{\BP}{\bm{P}}
\newcommand{\BQ}{\bm{Q}}

\newcommand{\BV}{\bm{V}}
\newcommand{\BW}{\bm{W}}
\newcommand{\BX}{\bm{X}}

\newcommand{\Balpha}{\bm{\alpha}}

\newcommand{\Bkappa}{\bm{\kappa}}

\newcommand{\loss}{\mathcal{L}}

\newcommand{\R}{\mathbb{R}}

\newcommand{\cutsectionup}{\vspace*{-0.15in}}
\newcommand{\cutsectiondown}{\vspace*{-0.12in}}
\newcommand{\cutsubsectionup}{\vspace*{-0.1in}}
\newcommand{\cutsubsectiondown}{\vspace*{-0.07in}}

\newcommand\modelname[0]{FPS-T}

\title{Curve Your Attention: Mixed-Curvature Transformers for Graph Representation Learning}

\author{%
  Sungjun Cho$^1$ \quad Seunghyuk Cho$^2$\thanks{Work done during an internship at LG AI Research.} \quad Sungwoo Park$^1$ \quad Hankook Lee$^1$\\
  \textbf{Honglak Lee}$^{1}$ \quad \textbf{Moontae Lee}$^{1,3}$\\
  $^1$LG AI Research \quad $^2$POSTECH \quad $^3$University of Illinois Chicago\\
}

\begin{document}

\maketitle

\begin{abstract}
  Real-world graphs naturally exhibit hierarchical or cyclical structures that are unfit for the typical Euclidean space. While there exist graph neural networks that leverage hyperbolic or spherical spaces to learn representations that embed such structures more accurately, these methods are confined under the message-passing paradigm, making the models vulnerable against side-effects such as oversmoothing and oversquashing. More recent work have proposed global attention-based graph Transformers that can easily model long-range interactions, but their extensions towards non-Euclidean geometry are yet unexplored. To bridge this gap, we propose Fully Product-Stereographic Transformer, a generalization of Transformers towards operating entirely on the product of constant curvature spaces. When combined with tokenized graph Transformers, our model can learn the curvature appropriate for the input graph in an end-to-end fashion, without the need of additional tuning on different curvature initializations. We also provide a kernelized approach to non-Euclidean attention, which enables our model to run in time and memory cost linear to the number of nodes and edges while respecting the underlying geometry. Experiments on graph reconstruction and node classification demonstrate the benefits of generalizing Transformers to the non-Euclidean domain.
\end{abstract}

\section{Introduction}
\cutsubsectiondown

Learning from graph-structured data is a challenging task in machine learning, with various downstream applications that involve modeling individual entities and relational interactions among them~\cite{(citation)sen2008collective, (power)watts1998collective, (web-edu)gleich2004fast}. A dominant line of work consists of graph convolutional networks (GCNs) that aggregate features across graph neighbors through \textit{message-passing}~\cite{(message-passing)gilmer2017neural, (GCN)kipf2016semi, (GAT)velivckovic2017graph, (SGC)wu2019simplifying, (SAGE)hamilton2017inductive}. While most GCNs learn features that lie on the typical Euclidean space with zero curvature, real-world graphs often comprise of complex structures such as hierarchical trees and cycles that Euclidean space requires excessive dimensions to accurately embed
~\cite{(hyperbolic_embedding5)pmlr-v80-sala18a}. In response, the graph learning community has developed generalizations of GCNs to spaces with non-zero curvature such as hyperbolic, spherical, or mixed-curvature spaces with both negative and positive curvatures~\cite{(HGCN)chami2019hyperbolic, (HGNN)liu2019hyperbolic, (HAT)zhang2021hyperbolic, (kappa-GCN)bachmann2020constant, (QGCN)xiong2022pseudo}.

Unfortunately, non-Euclidean GCNs are not immune to harmful side-effects of message-passing such as oversmoothing~\cite{(oversmoothing1)oono2019graph, (oversmoothing2)cai2020note, (oversmoothing)yang2022hrcf} and oversquashing~\cite{(Balanced-Forman)topping2021understanding, (oversquashing)alon2020bottleneck}. These drawbacks make it difficult to stack GCN layers towards large depths, limiting its expressive power~\cite{(expressivity)feng2022powerful, (expressivity)maron2019provably} as well as predictive performance on tasks that require long-range interactions to solve~\cite{(LRG)dwivedi2022long, liu2021non}.
To cope with such limitations, recent works have instead proposed Transformer-based graph encoders that can easily exchange information across long-range distances through global self-attention~\cite{(TokenGT)kim2022pure, (Graphormer)ying2021transformers, (GT)dwivedi2020generalization, (SAN)kreuzer2021rethinking}. However, existing graph Transformers are still confined within the Euclidean regime, and their extensions towards non-Euclidean geometry has not yet been studied.

In this paper, we bridge this gap by generalizing the Transformer architecture~\cite{(Transformer)vaswani2017attention} towards non-Euclidean spaces with learnable curvatures. Specifically, we endow each attention head a stereographic model~\cite{(kappa-GCN)bachmann2020constant} that can universally represent Euclidean, hyperbolic, and spherical spaces (Figure~\ref{fig:stereographic}). 
We generalize each operation of the Transformer architecture to inputs on the product-stereographic model, all of which are end-to-end differentiable with respect to the sectional curvatures, thereby allowing the model to jointly train embeddings as well as the underlying curvature. 
The resulting model, which we name as \textbf{Fully Product-Stereographic Transformer (\modelname{})}, takes advantage of both non-Euclidean geometry and long-range interactions. 
We empirically show that the learnable sectional curvature of \modelname{} successfully converges to the geometry of the input graph, leading to better predictive performance and parameter efficiency in graph reconstruction and node classification compared to its Euclidean counterpart. To the best of our knowledge, our work is the first to propose a natural generalization of Transformers to non-Euclidean spaces. 



We summarize our core contributions as follows:
\begin{itemize}
    \item We propose \modelname{}, a generalization of Transformer towards operating entirely on the product-stereographic model with curvatures that are learnable in an end-to-end fashion.
    \item For graph representation learning, we integrate \modelname{} with Tokenized Graph Transformer~\cite{(TokenGT)kim2022pure}, and develop a kernelized approximation of non-Euclidean attention to reduce the computational cost to linear in number of nodes and edges.
    \item Experiments on graph reconstruction and node classification with real-world graphs demonstrate the benefits of \modelname{} such as better parameter efficiency and downstream performance.
\end{itemize}

\definecolor{c0}{HTML}{1f77b4}
\definecolor{c1}{HTML}{ff7f0e}
\definecolor{c2}{HTML}{2ca02c}
\begin{figure}[t!]
    \centering
    \begin{tikzpicture}
        \node[anchor=south west] at (0, 0) {\includegraphics[width=.9\linewidth]{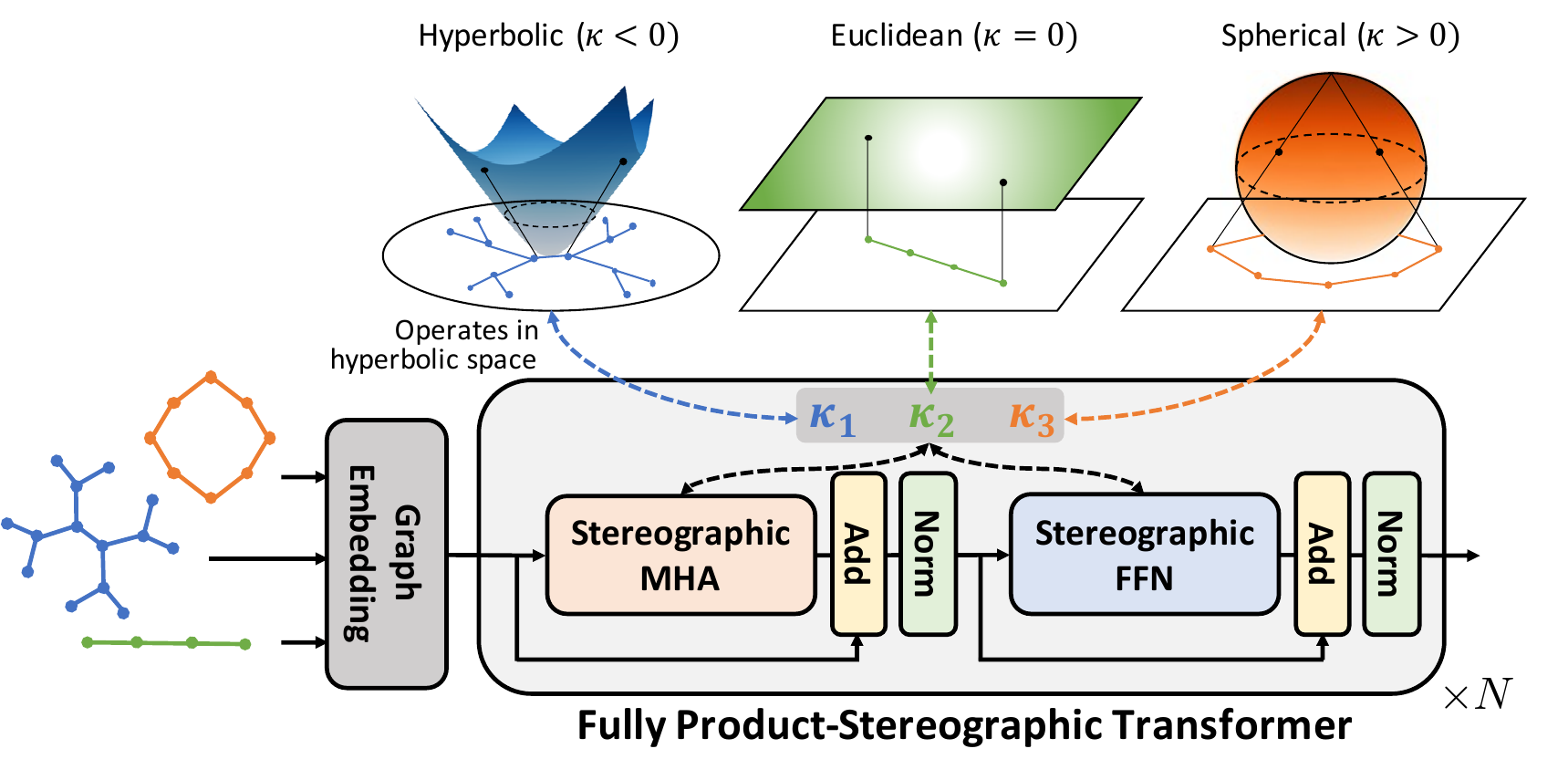}};
    \end{tikzpicture}
    \vspace{-.1in}
    \caption{
    Illustration of our proposed \modelname{} architecture.
    Well-known constant curvature spaces can be projected to the stereographic model, with a common chart map isomorphic to the $d$-dimensional Euclidean space.
    Each space can efficiently embed different types of graphs (\textit{e.g.}, \textcolor{c0}{trees in hyperbolic space}, \textcolor{c2}{lines in Euclidean space}, and \textcolor{c1}{cycles in spherical space}).
    In \modelname{}, each layer chooses a set of curvatures that fits the input graph by changing the sign of the curvature $\kappa$ in a differentiable manner.
    }
    \label{fig:stereographic}
    \vspace{-.1in}
\end{figure}

\section{Related Work}

\paragraph{Non-Euclidean graph representations.}
Non-Euclidean spaces are known to well-preserve specific types of graph structure where Euclidean space fails.
Especially, non-Euclidean spaces with constant sectional curvature, \textit{e.g.}, hyperbolic and spherical spaces, are widely used in graph representation learning due to its tractable operations.
Hyperbolic spaces are capable of efficiently embedding complex hierarchical structures in graphs~\cite{(hyperbolic_embedding1)nickel2018learning, (hyperbolic_embedding2)NIPS2017_59dfa2df, (hyperbolic_embedding3)ganea2018hyperbolic, (hyperbolic_embedding4)krioukov2010hyperbolic, (hyperbolic_embedding5)pmlr-v80-sala18a}.
Graphs with cyclic structures are well-suited for spherical spaces~\cite{(spherical_embedding1)6787114, (spherical_embedding2)grattarola2019change}.
Riemannian manifolds with varying curvature and constant sign are also proposed for graph encoding~\cite{(tractable)cruceru2021computationally}.
However, Riemannian manifolds where the sign of the curvature is fixed are not a good choice for more complex graphs that exhibit both hierarchy and cycles.
Instead, the product of constant-curvature spaces~\cite{(mixed_curvature)gu2018learning}, heterogeneous manifolds~\cite{(heterogeneous)giovanni2022heterogeneous}, and pseudo-Riemannian manifolds~\cite{(pseudo_Riemannian)law2020ultrahyperbolic} are found to be well-suited for learning representations of such complex graphs.

Message passing GCNs also benefit from considering a non-Euclidean representation space.
Hyperbolic GCNs are known to outperform Euclidean counterparts in various tasks on hierarchical graphs such as citation networks~\cite{(HGCN)chami2019hyperbolic, (HAT)zhang2021hyperbolic, (geomgcn)Pei2020Geom-GCN:} and molecules~\cite{(HGCN)chami2019hyperbolic, (HGNN)liu2019hyperbolic}.
Deepsphere~\cite{(deepsphere)Defferrard2020DeepSphere:} also adopted the spherical space to GCNs with applications such as 3D object and earth climate modeling.
To take the advantage of multiple spaces, \cite{(GIL)zhu2020graph} proposed a hybrid architecture that fuses Euclidean and hyperbolic graph representations together. \cite{(FMGNN)deng2023fmgnn} similarly proposed modeling interactions between three constant-curvature spaces (\textit{i.e.}, Euclidean, hyperbolic, and spherical). 
To allow smooth connections between the three constant-curvature spaces, \cite{(kappa-GCN)bachmann2020constant} proposed a model of constant-curvature space called the stereographic model, on which geometric operations such as distances and inner products are differentiable at all curvature values including zero.
Incorporating pseudo-Riemannian manifolds with the GCN architecture also showed promising results~\cite{(QGCN)xiong2022pseudo}, but its performance is sensitive to the time dimension of the manifold, which requires extensive hyperparameter tuning.

Overall, GCNs achieve great predictive performance in homophilic graphs where connected nodes share the same features, but they tend to fail in hetereophilic graphs, as stacking up GCN layers to capture message passing between distant nodes induces oversmoothing~\cite{(oversmoothing1)oono2019graph, (oversmoothing2)cai2020note} and oversquashing~\cite{(Balanced-Forman)topping2021understanding}. 
To relieve this architectural limitation while utilizing non-Euclidean geometrical priors, we instead develop a Transformer-based graph encoder that operates on the steregraphic model to learn graph representations.

\paragraph{Graph Transformers.} Inspired by huge success of Transformers in NLP and CV~\cite{(BERT)devlin2018bert, (GPT-3)brown2020language, (ViT)dosovitskiy2020image}, there exist various work that extend Transformers for encoding graphs with edge connectivities that are neither sequential nor grid-like. Graph Transformer~\cite{(GT)dwivedi2020generalization} and Spectral Attention Network~\cite{(SAN)kreuzer2021rethinking} were the first pioneers to explore this direction by replacing sinusoidal positional encodings widely used in NLP with Laplacian eigenvectors of the input graph. Graphormer~\cite{(Graphormer)ying2021transformers} then proposed utilizing edge connectivities by using shortest-path distances as an attention-bias, showing state-of-the-art performance on molecular property prediction. TokenGT proposed a tokenization technique that views each graph as a sequence of nodes and edges. Unlike other methods, TokenGT allows straightforward integration of engineering techniques of pure Transformers such as linearized attention~\cite{(linearized_attention)katharopoulos2020transformers}, while enjoying theoretical expressivity that surpasses that of message-passing GCNs.

Nonetheless, existing Transformer architectures for graphs are yet confined within the Euclidean domain, making them unable to precisely embed graphs onto the feature space similar to geometric GCNs. While Hyperbolic Attention Network~\cite{(hyp-attn)gulcehre2018hyperbolic} proposed an attention mechanism that operates on hyperbolic space, 
its distance-based attention imposes a computational cost quadratic to the graph size and the geometry is limited to hyperbolic space.
Instead, we generalize the representation space of Transformer to stereographic model and integrate with TokenGT, which can cover more various types of graphs.
We also linearize the attention mechanism on the stereographic model similar to ~\cite{(linearized_attention)katharopoulos2020transformers}, which allows our final model to run in cost linear to the number of nodes and edges.

\section{Preliminaries}
\cutsubsectiondown

In this section, we first explain the concepts related to our main geometrical tool, the product-stereographic model~\cite{(kappa-GCN)bachmann2020constant}.
We then briefly discuss multi-head attention, the main driving force of the Transformer~\cite{(Transformer)vaswani2017attention} model.

\cutsubsectionup
\subsection{Product-Stereographic Model}\label{sec:stereographic_background}
\cutsubsectiondown

\paragraph{Riemannian manifolds.}
A Riemannian manifold is consisted of a smooth manifold $\mathcal{M}$ and a metric tensor $g$.
Each point $\Bx$ on the manifold $\mathcal{M}$ defines a tangent space $\mathcal{T}_{\Bx} \mathcal{M}$, which is a collection of all vectors that are tangent to $\Bx$, also called the tangent vector.
The metric tensor $g: \mathcal{M} \rightarrow \R^{n \times n}$ assigns a positive-definite matrix to each point $\Bx$, which defines its inner product $\langle \cdot, \cdot \rangle_{\Bx}: \mathcal{T}_{\Bx} \mathcal{M} \times \mathcal{T}_{\Bx} \mathcal{M} \rightarrow \R$ as $\Bv_1^T g(\Bx)\Bv_2$ where $\Bv_1, \Bv_2 \in \mathcal{T}_{\Bx}\mathcal{M}$ are the tangent vectors of $\Bx$.

The metric tensor is used to define geometrical properties and operations of the Riemannian manifold.
Geodesic $\gamma$ is the shortest curve between two points $\Bx, \By \in \mathcal{M}$ and its distance can be computed as $d_{\mathcal{M}}(\Bx, \By) = \int_0^1 \langle \dot{\gamma}(t), \dot{\gamma}(t) \rangle_{\gamma(t)} dt$, where $\gamma: [0, 1] \rightarrow \mathcal{M}$ is a unit-speed curve satisfying $\gamma(0) = \Bx$ and $\gamma(1) = \By$.


We can move the point $\Bx \in \mathcal{M}$ along a tangent vector $\Bv \in \mathcal{T}_{\Bx} \mathcal{M}$ using exponential map $\exp_{\Bx}: \mathcal{T}_{\Bx} \mathcal{M} \rightarrow \mathcal{M}$ which is defined as $\exp_{\Bx}(\Bv) = \gamma(1)$ where $\gamma$ is a geodesic and $\gamma(0) = \Bx, \dot{\gamma(0)} = \Bv$.
The logarithmic map $\log_{\Bx}: \mathcal{M} \rightarrow \mathcal{T}_{\Bx} \mathcal{M}$ is the inverse of $\exp_{\Bx}$.
A tangent vector $\Bv \in \mathcal{T}_{\Bx} \mathcal{M}$ can be transferred along a geodesic from $\Bx$ to $\By$ using parallel transport $\mathrm{PT}_{\Bx \rightarrow \By}: \mathcal{T}_{\Bx} \mathcal{M} \rightarrow \mathcal{T}_{\By} \mathcal{M}$.

Note that the product of Riemannian manifolds is also a Riemannian manifold.
A point on the product Riemannian manifold $\Bx \in \otimes_{i=1}^n \mathcal{M}_i$ is consisted of the parts from each Riemannian manifold $\mathcal{M}_i$ which is written as $\Bx = \Vert_{i=1}^n \Bx_i$, where $\Bx_i \in \mathcal{M}_i$ and $\Vert$ is the concatenation operation.
The distance between $\Bx, \By \in \otimes_{i=1}^n \mathcal{M}_i$ is calculated as $\sqrt{\sum_{i=1}^n d_{\mathcal{M}_i}^2(\Bx_i, \By_i)}$.
Other operations such as exponential map, logarithmic map, and parallel transport are applied manifold-wise. For example, $\exp_{\Bx}(\Bv) = \Vert_{i=1}^n \exp_{\Bx_i}(\Bv_i)$, where $\Bv = \Vert_{i=1}^n \Bv_i$ and $\Bv_i \in \mathcal{T}_{\Bx_i} \mathcal{M}_i$.

\paragraph{Constant-curvature spaces.}

Curvature is an important geometrical property used to characterize Riemannian manifolds.
One of the widely-used curvatures to explain Riemannian manifolds is the sectional curvature: given two linearly independent tangent vector fields $U, V \in \mathfrak{X}(\mathcal{M})$, the sectional curvature $K(U, V)$ is computed as $K(U, V) = \frac{\langle R(U, V)V, U \rangle}{\langle U, U \rangle \langle V, V \rangle - \langle U, V \rangle^2}$,
where $R(\cdot, \cdot): \mathfrak{X}(\mathcal{M}) \times \mathfrak{X}(\mathcal{M}) \times \mathfrak{X}(\mathcal{M}) \rightarrow \mathfrak{X}(\mathcal{M})$ is a Riemannian curvature tensor.
The sectional curvature measures the divergence between the geodesics starting with the tangent vector fields $U, V$ for each point of the manifold.
For the positive or negative sectional curvatures, geodesics become closer or farther than the zero-curvature case, respectively.


Throughout this paper, we refer to a space of a constant sectional curvature as a constant-curvature space.
For example, the Euclidean space is the special case of the constant-curvature space with zero curvature.
For positive and negative cases, we call the spaces as hyperbolic and spherical spaces, respectively.

\paragraph{Stereographic models.}
A $d$-dimensional stereographic model $\mathfrak{st}^d_{\kappa}$ is a constant-curvature space with curvature value $\kappa$.
One attractive property of the stereographic model is that the operations such as distance, exponential map, logarithmic map, and parallel transport are differentiable at any curvature value $\kappa$, including $\kappa=0$.
This enables the stereographic model to learn the curvature value $\kappa$ without any constraint.

The manifold of the stereographic model $\mathfrak{st}_{\kappa}^d$ is $\{\Bx \in \R^d \vert -\kappa \Vert \Bx \Vert^2 < 1 \}$.
The metric tensor is defined as $g^{\kappa}(\Bx) = \frac{4}{1 + \kappa \Vert \Bx \Vert^2}\BI =: (\lambda_{\Bx}^{\kappa})^2 \BI$, where $\lambda_{\Bx}^\kappa$ is known as the conformal factor.
The mobius addition between two points $\Bx, \By \in \mathfrak{st}_\kappa^d$ is computed as $\Bx \oplus_{\kappa} \By = \frac{(1 - 2\kappa \Bx^T \By - \kappa \Vert \By \Vert^2)\Bx + (1 + \kappa \Vert \Bx \Vert^2)\By}{1-2\kappa \Bx^T \By + \kappa^2 \Vert \Bx \Vert^2 \Vert \By \Vert^2}$.
Based on mobius addition, we can derive other geometric operations as
\autoref{tab:stereographic_ops} in Appendix~\ref{apx:stereographic}.
The table also shows that when $\kappa$ converges to zero, the operations become equivalent to Euclidean space operations, so the stereographic model essentially recovers Euclidean geometry.

\subsection{Multi-Head Attention}
\cutsubsectiondown

In vanilla Transformer~\cite{(Transformer)vaswani2017attention}, each attention block contains multiple attention heads, each taking a sequence of token embeddings as input $\BX \in \R^{n \times d}$ with sequence length $n$ and feature dimension $d$. Three trainable linear weights $\BW^Q, \BW^K, \BV^V \in \R^{d \times d'}$ first map each token embedding into queries $\BQ$, keys $\BK$, and values $\BV$ with head-dimension $d'$, respectively. Then, the attention score matrix is computed by scaled Euclidean dot-product between $\BQ$ and $\BK$, followed by row-wise softmax activation $\sigma(\cdot)$. The attention score matrix is then multiplied to value $\BV$, returning contextualized token embeddings. The overall procedure can be written as
\begin{gather}
    \BQ = \BX \BW^Q,\;\; \BK = \BX \BW^K,\;\; \BV = \BX \BW^V,\\
    \text{Attn}(\BX) = \sigma\left(\dfrac{\BQ\BK^T}{\sqrt{d'}}\right) \BV.
\end{gather}
The output from multiple attention heads are concatenated together, then processed through a feed-forward layer before proceeding to the next Transformer block.

\section{Fully Product-Stereographic Transformer}

Here, we describe the inner wirings of our proposed method. We generalize each operation in Transformer to the product-stereographic model, together forming a geometric Transformer architecture that operates entirely within the stereographic model.

\subsection{Stereographic Neural Networks}

We first introduce the stereographic analogies of the Euclidean neural networks such as the linear layer, activation, layer normalization, and logit functions.
We denote the product-stereographic model $\otimes_{i=1}^H \mathfrak{st}_{\kappa_i}^d$ as $\mathfrak{st}_{\otimes \Bkappa}^d$, where $\Bkappa = (\kappa_1, \dots, \kappa_H)$ is the ordered set of curvatures of $d$-dimensional component spaces within a Transformer block with $H$ attention heads.
We also use the superscript $\otimes \Bkappa$ to denote Riemannian operations on product-stereographic model that decompose representations into equal parts, apply the operation, then concatenate back to the product space (\textit{e.g.}, if $\Bv = [v_1, \dots, v_H]$, then $\exp_{\Bzero}^{\otimes \Bkappa}(\Bv) \coloneqq \Vert_{i=1}^H \exp_{\Bzero}^{\kappa_i}(v_i)$).



\paragraph{Stereographic linear layer, activation, and layer normalization.} 
Given a Euclidean neural network $f$, we can define its stereographic counterpart as $\exp_{\Bzero}^{\otimes\Bkappa}\left( f\left(\log_{\Bzero}^{\otimes\Bkappa}(\BX)\right)\right)$.
The stereographic linear layer $\textrm{Linear}_{\otimes \Bkappa}(\BX; \BW)$ is thus defined by setting $f$ as the Euclidean linear layer $f(\BX; \BW) = \BX\BW$.
The same approach can be used for any Euclidean activation function $f_{\textrm{act}}$ (\textit{e.g.}, \textrm{ReLU}, \textrm{Tanh}, \textrm{ELU}, and \textrm{Sigmoid}), from which we obtain stereographic activation functions.
Stereographic layer normalization $\textrm{LN}_{\otimes \Bkappa}$ is defined in the same manner.





\paragraph{Stereographic logits.} 

Suppose that $\Bx \in \mathfrak{st}_{\Bkappa}^d$ is a stereographic embedding retrieved from the last transformer layer.
For prediction tasks such as node classification, we need to compute the probability that the node with embedding $\Bx$ belongs to class $c$.
Inspired by logistic regression of Euclidean space, \cite{(kappa-GCN)bachmann2020constant} proposes its stereographic variant as:
\begin{equation}
    p(y = c \mid \Bx) \propto \exp \left( \textrm{sign}(\langle -\Bp_c \oplus_{\Bkappa} \Bx, \Ba_c \rangle )\Vert \Ba_c \Vert_{\Bp_c} d_{\Bkappa}(\Bx, H_{\Ba_c, \Bp_c}) \right),
\end{equation}
where $H_{\Ba_c, \Bp_c} = \{ \Bx \in \mathfrak{st}_{\Bkappa}^d \mid \langle -\Bp_c \oplus_{\Bkappa} \Bx, \Ba_c \rangle = 0 \}$ is a hyperplane formed by $\Ba_c \in \mathcal{T}_{\Bp_c} \mathfrak{st}_{\Bkappa}^d$ and $\Bp_c \in \mathfrak{st}_{\Bkappa}^d$. 
For a stereographic model $\mathfrak{st}_\kappa^d$, the distance between $\Bx \in \mathfrak{st}_\kappa^d$ and the hyperplane $H_{\Ba, \Bp}$ is derived as:
\begin{equation}
    d_\kappa(\Bx, H_{\Ba, \Bp}) = \sin^{-1}_{\kappa * \vert \kappa \vert} \left( \frac{2 \vert \langle -\Bp \oplus_{\kappa} \Bx, \Ba \rangle \vert}{(1 + \kappa \Vert \langle -\Bp \oplus_{\kappa} \Bx, \Ba \rangle \Vert^2)\Vert \Ba \Vert} \right).
\end{equation}
This distance function can be easily extended to the product-stereographic model as mentioned in Section \ref{sec:stereographic_background}.
The parameters $\Ba, \Bp$ that define the hyperplane are learned together with the model parameters during the training phase.

\subsection{Stereographic Multi-Head Attention}

\begin{figure}[t!]
    \centering
    \begin{subfigure}[t]{.48\textwidth}
    \centering
    \includegraphics[width=\textwidth]{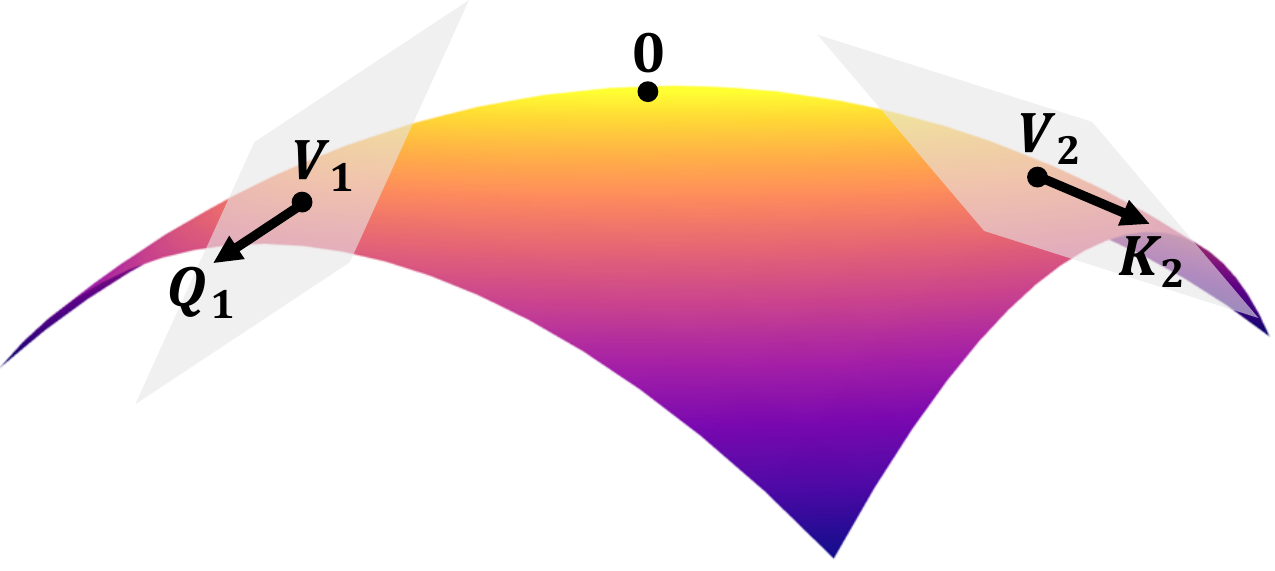}
    \caption{QKV mapping}
    \end{subfigure}
    \hspace{0mm}
    \begin{subfigure}[t]{.48\textwidth}
    \centering
    \includegraphics[width=\textwidth]{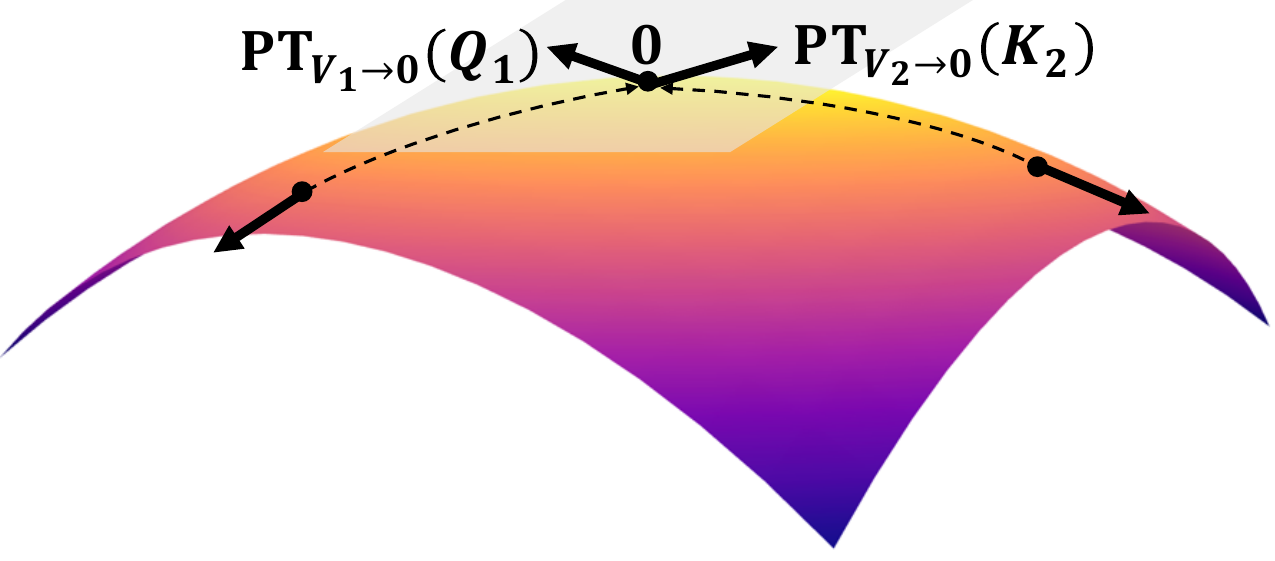}
    \caption{Parallel transport and inner product}
    \end{subfigure}
    \caption{Illustration of our attention mechanism on the non-Euclidean space. \modelname{} considers each value-vector as a point that resides on the stereographic model, and query/key-vectors as tangent vectors on the corresponding tangent spaces. All query/key-vectors are parallel-transported to the origin prior to dot-product attention, thereby taking the given geometry into account.}
    \label{fig:architecture}
\end{figure}

Using the stereographic operations and neural networks above, we propose a multi-head attention mechanism under product-stereographic models. The key intuition is that each $h$-th attention head operates on the $\kappa_h$-stereographic space.
Given a sequence of $n$ product-stereographic embeddings $\BX \in \mathfrak{st}_\kappa^{n \times d}$, the attention head with curvature $\kappa$ first obtains values using the stereographic linear layer.
For queries and keys, it maps each stereographic embedding to the tangent space of the values as:
\begin{align}
    \BQ = \BX\BW^Q \in \mathcal{T}_{\BV} \mathfrak{st}_\kappa^{n \times d'},\;\;
    \BK = \BX\BW^K \in \mathcal{T}_{\BV} \mathfrak{st}_\kappa^{n \times d'},\;\;
    \BV = \mathrm{Linear}_{\kappa} (\BX; \BW^V) \in \mathfrak{st}_\kappa^{n \times d'},
\end{align}
where $\BW^Q, \BW^K \in \R^{d \times d'}$ are the query/key weight matrices, and $\BW^V \in \R^{d \times d'}$ is the weight matrix for values.
Note that the constraint of the tangent space of the stereographic model $\mathfrak{st}_\kappa^d$ is the same at all the points as $\R^d$.

Then, the attention-score between the $i$ th query $\BQ_i$ and $j$ th key $\BK_j$ is computed by parallel transporting the vectors to the origin, and taking the Riemannian inner product at the origin as 
\begin{equation}\label{eq:st_attention}
    \alpha_{ij} = \langle \mathrm{PT}_{\BV_i \to \Bzero}(\BQ_i), \mathrm{PT}_{\BV_j \to \Bzero}(\BK_j) \rangle_{\Bzero}. 
\end{equation}
\autoref{fig:architecture} illustrates the geometric attention mechanism.
Because the metric tensor of the origin of the stereographic model is simply $4\BI$ with identity matrix $\BI$, the Riemannian inner product becomes equivalent to the Euclidean inner product at the origin.

Finally, we aggregate the values based on the attention scores using the Einstein midpoint~\cite{(kappa-GCN)bachmann2020constant} as
\begin{equation}
\label{eq:st_aggregation}
    \textrm{Aggregate}_\kappa \left(\BV, \Balpha\right)_i \coloneqq \dfrac{1}{2} \otimes_{\kappa} \left(\sum_{j=1}^n \dfrac{\alpha_{ij} \lambda_{\BV_j}^\kappa}{\sum_{k=1}^n \alpha_{ik} (\lambda_{\BV_k}^\kappa - 1)}\BV_j \right),
\end{equation}
with conformal factors $\lambda_{\BV_i}^\kappa$ at point $\BV_i \in \mathfrak{st}_{\kappa}^{d'}$.
By concatenating the aggregated results from each attention head, the final outcome of product-stereographic multi-head attention is
\begin{equation}
    \textrm{MHA}_{\otimes \Bkappa}(\BX) = \Vert_{h=1}^H \textrm{Aggregate}_{\kappa_h}(\BV^h, \Balpha^h) \in \otimes_{h=1}^H \mathfrak{st}_{\kappa_h}^{n \times d},
\end{equation}
where $\kappa_h$ denotes the curvature of the $h$-th attention head.

\subsection{Wrap-up}

For completeness, we fill in the gap on how intermediate steps such as skip-connection are generalized towards non-zero curvatures, and how representations are processed between Transformer layers with distinct curvatures.
First, recall that vanilla Transformer utilizes residual connections and Layer normalization to mitigate vanishing gradients and induce better convergence~\cite{(Transformer)vaswani2017attention}. To apply these operations on representations in the product-stereographic space, we switch to
\begin{gather}
    \BX_l = \textrm{MHA}_{\otimes\Bkappa} (\textrm{LN}_{\otimes\Bkappa}(\BX^{\textrm{in}}_{l})) \oplus_{\kappa} \BX^{\textrm{in}}_{l}\label{eq:mha} \\
    \BX^{\textrm{out}}_l = \textrm{FFN}_{\otimes\Bkappa} (\textrm{LN}_{\otimes\Bkappa}(\BX_l)) \oplus_{\kappa} \BX_l. \label{eq:ffn}
\end{gather}
Note that while each attention head in stereographic multi-head attention operates on each stereographic model independently, the product-stereographic feed-forward network $\textrm{FFN}_{\otimes\Bkappa}$, for which we use two stereograhpic linear layers with an activation in between, fuses representations from distinct geometries and performs interactions between different steregraphic models similarly to previous work~\cite{(GIL)zhu2020graph, (FMGNN)deng2023fmgnn}.

Furthermore, note that each $l$-th Transformer layer operates on a distinct product-stereographic space $\mathfrak{st}_{\otimes \Bkappa^l}^d$ where $\Bkappa^l = (\kappa^l_1, \dots, \kappa^l_H)$ together forms the geometric signature of the layer. 
For consistency, we assume that the input embeddings are on the product-stereographic model of the first layer (\textit{i.e.}, $\mathfrak{st}_{\otimes \Bkappa^1}^d$).
In case of classification tasks where logits are computed, the product-stereographic logit layer operates on the last set of curvatures (\textit{i.e.}, $\mathfrak{st}_{\otimes \Bkappa^L}^d$ where $L$ denotes the number of Transformer layers). In between layers, representations are translated from $\mathfrak{st}_{\otimes \Bkappa^l}^d$ to $\mathfrak{st}_{\otimes \Bkappa^{l+1}}^d$ by assuming a shared tangent space at the origin (\textit{i.e.}, $\BX_{l+1}^{\textrm{in}} = (\exp_{\Bzero}^{\otimes \Bkappa_{l+1}} \circ \log_{\Bzero}^{\otimes\Bkappa_l})(\BX_{l}^{\textrm{out}})$).

Altogether, it is straightforward to find that \textbf{\modelname{} becomes equivalent to the original Transformer as all $\Bkappa$ approaches 0}, but it possesses the capability to deviate itself away from Euclidean geometry if it leads to better optimization.
For all experiments, we initialize all curvatures as zero to demonstrate the practicality of our method by not requiring extensive hyperparameter tuning over different combinations of curvatures.


\subsection{Extension to Graph Transformer}

In order to learn graph-structured data with \modelname{}, we borrow the tokenization technique proposed by TokenGT~\cite{(TokenGT)kim2022pure}. Let graph $\mathcal{G} = (\mathcal{V}, \mathcal{E})$ be an input graph with $N$ nodes in node-set $\mathcal{V}$, $M$ edges in edge-set $\mathcal{E}$, and respective features $\BX^\mathcal{V} \in \R^{N \times d}$, $\BX^\mathcal{E} \in \R^{M \times d}$. Then, we tokenize the graph into a sequence $\BX = [\BX^{\mathcal{V}}, \BX^{\mathcal{E}}] \in \R^{(N+M) \times d}$ by treating each node and edge as an independent token, and augment the tokens with 1) node identifiers that serve as positional encoding and 2) type identifiers that allows the model to distinguish between node- and edge-tokens. TokenGT feeds this sequence into a pure Euclidean Transformer, an approach proven to pass the 2-dimensional Weisfeiler-Lehman (2-WL) graph isomorphism test and surpass the theoretical expressivity of message-passing GCNs~\cite{(TokenGT)kim2022pure, (expressivity)maron2019provably}. More details on the tokenization procedure can be found in \autoref{apx:architecture}.

In our work, we encode the input sequence through \modelname{} instead, such that nodes and edges exchange information globally on the product-stereographic space. As augmented tokens $\BX$ are Euclidean vectors, we assume each token lies within the tangent space at the origin of the product-stereographic model of the first layer $\mathcal{T}_{\Bzero} \mathfrak{st}_{\otimes \Bkappa^1}^{d'} \cong \R^{H \times d'}$, where $\vert \Bkappa^1 \vert = H$ and $Hd' = d$. 
Therefore, apply exponential mapping on the tokens to place them on the product-stereographic model via $\exp_{\Bzero}^{\otimes \Bkappa^1}(\BX)$, the output of which is forwarded through \modelname{}.


\subsection{Cost Linearization of Stereographic Attention}

One drawback of the graph tokenization method above is its computational cost that becomes intractable when encoding large graphs. As computing the attention score matrix takes time and memory quadratic to the sequence length, a graph with $N$ nodes and $M$ edges incurs an asymptotic cost of $\mathcal{O}((N+M)^2)$, which can be $\mathcal{O}(N^4)$ for dense graphs. 
Fortunately, there exist various advancements used to make Transformers more efficient~\cite{(ET_survey)tay2022efficient, (reformer)kitaev2020reformer, (performer)choromanski2020rethinking, (linformer)wang2020linformer, (nystromformer)xiong2021nystromformer, (SBM-Transformer)cho2022transformers}.


In previous work~\cite{(linearized_attention)katharopoulos2020transformers}, it is shown that the Euclidean attention score $\langle \BQ_i, \BK_j \rangle$ can be approximated with the product of kernel function $\phi(\BQ_i)\phi(\BK_j)$, where $\phi(\BX) = \textrm{ELU}(\BX) + 1$.
For stereographic attention (\autoref{eq:st_attention}), computing dot-products on the tangent space of the origin allows us to extend this kernelization to \modelname{}. 
Let $\tilde{\BQ}_i = \textrm{PT}_{\BV_i \rightarrow \Bzero}(\BQ_i)$ and $\tilde{\BK}_j = \textrm{PT}_{\BV_j \rightarrow \Bzero}(\BK_j)$ be the tangent vectors on the origin prior to taking the dot-product.
By applying the kernelization to stereographic attention, we can rewrite the stereographic aggregation (\autoref{eq:st_aggregation}) as:
\begin{align}
    \dfrac{1}{2} \otimes_{\kappa} \left(\sum_{j=1}^n \dfrac{\langle \tilde{\BQ}_i, \tilde{\BK}_j \rangle_{\Bzero} \lambda_{\BV_j}^\kappa}{\sum_{k=1}^n \langle \tilde{\BQ}_i, \tilde{\BK}_k \rangle_{\Bzero} (\lambda_{\BV_k}^\kappa - 1)}\BV_j \right) 
    \approx \dfrac{1}{2} \otimes_{\kappa} \left[\phi(\tilde{\BQ})\left(\phi'(\tilde{\BK})^T \tilde{\BV}\right)\right]_i,
\end{align}
where $\phi'(\BK)_i = \phi(\BK)_i (\lambda^\kappa_{\BV_i} - 1)$ and $\tilde{\BV}_i = \frac{\lambda^\kappa_{\BV_i}}{\lambda^\kappa_{\BV_i} - 1} \BV_i$.

This approximation enables \modelname{} to encode graphs with $\mathcal{O}(N+M)$ cost, which matches the complexity of message-passing GCNs~\cite{(asymptotic_cost)wu2020comprehensive}, while taking the non-Euclidean geometry into account. 
In our experiments, we use the linearized \modelname{} 
and find that this approach performs well in practice.

\section{Experiments}

We empirically test the performance of \modelname{} on graph reconstruction and node classification tasks.
We compare the performance to existing baselines such as message passing-based Euclidean (GCN~\cite{(GCN)kipf2016semi}, GAT~\cite{(GAT)velivckovic2017graph}, SAGE~\cite{(SAGE)hamilton2017inductive}, SGC~\cite{(SGC)wu2019simplifying}), hyperbolic (HGCN~\cite{(HGCN)chami2019hyperbolic}, HGNN~\cite{(HGNN)liu2019hyperbolic}, HAT~\cite{(HAT)zhang2021hyperbolic}), and mixed-curvature ($\kappa$-GCN~\cite{(kappa-GCN)bachmann2020constant}, $\mathcal{Q}$-GCN~\cite{(QGCN)xiong2022pseudo}) GCNs.
We also add TokenGT as our baseline, which is equivalent to \modelname{} with fixed zero curvatures.
Our model is implemented using PyTorch~\cite{(pytorch)paszke2019pytorch}, PyTorch Geometric~\cite{(pytorch_geometric)fey2019fast}, and Geoopt~\cite{(geoopt)geoopt2020kochurov}. All experiments are run on NVIDIA A100 GPUs.

\subsection{Graph Reconstruction}

\paragraph{Datasets.} We experiment graph reconstruction of four different real-world networks. Web-Edu~\cite{(web-edu)gleich2004fast} is a web-page network under the \textit{.edu} domain connected with hyperlinks.
Power~\cite{(power)watts1998collective} is a network that models the electrical power grid in western US.
Bio-Worm~\cite{(bio-worm)cho2014wormnet} is a genetics network of the \textit{C. elegans} worm.
Facebook~\cite{(facebook)leskovec2012learning} is a social network. The detailed statistics of the datasets can be found in \autoref{apx:dataset_statistics}.

\begin{figure}
    \begin{minipage}[t]{.57\textwidth}
        \centering
        \vspace{0pt}
        \resizebox{\textwidth}{!}{\begin{tabular}{l|cccc}
            \toprule
            Dataset & Web-Edu & Power & Facebook & Bio-Worm\\
            Curvature & -0.63 & -0.28 & -0.08 & -0.03 \\ 
            \midrule
            \textsc{MLP} & 83.24{\scriptsize$\pm$1.32} & 83.89{\scriptsize$\pm$4.02} & 50.64{\scriptsize$\pm$15.12} & 73.34{\scriptsize$\pm$20.85} \\
            \textsc{GCN} & 79.95{\scriptsize$\pm$0.23} & 98.25{\scriptsize$\pm$0.02} & 78.99{\scriptsize$\pm$0.29} & 93.32{\scriptsize$\pm$1.06} \\
            \textsc{GAT} & 88.86{\scriptsize$\pm$0.36} & 99.03{\scriptsize$\pm$0.01} & 82.81{\scriptsize$\pm$0.25} & 97.76{\scriptsize$\pm$0.03} \\
            \textsc{SAGE} & 86.34{\scriptsize$\pm$0.31} & 97.58{\scriptsize$\pm$0.14} & 81.01{\scriptsize$\pm$0.26} & 96.86{\scriptsize$\pm$0.06} \\
            \textsc{SGC} & 78.78{\scriptsize$\pm$0.12} & 97.69{\scriptsize$\pm$0.05} & 74.69{\scriptsize$\pm$0.36} & 89.73{\scriptsize$\pm$0.59} \\
            \textsc{TokenGT} & 89.56{\scriptsize$\pm$0.03} & 99.08{\scriptsize$\pm$0.00} & 84.62{\scriptsize$\pm$0.13} & 97.75{\scriptsize$\pm$0.03} \\
            \midrule
            \textsc{HGCN} & 80.13{\scriptsize$\pm$0.31} & 96.82{\scriptsize$\pm$0.08} & 74.35{\scriptsize$\pm$5.39} & 86.96{\scriptsize$\pm$0.30} \\
            \textsc{HGNN} & 83.64{\scriptsize$\pm$0.26} & 97.85{\scriptsize$\pm$0.05} & 78.74{\scriptsize$\pm$0.58} & 90.97{\scriptsize$\pm$1.06} \\
            \textsc{HAT} & 90.21{\scriptsize$\pm$0.36} & 93.86{\scriptsize$\pm$0.34} & 80.09{\scriptsize$\pm$0.20} & 93.58{\scriptsize$\pm$0.42} \\
            \midrule
            \textsc{$\kappa$-GCN} & 55.34{\scriptsize$\pm$35.88} & 98.23{\scriptsize$\pm$0.09} & 20.80{\scriptsize$\pm$20.69} & 84.16{\scriptsize$\pm$13.67} \\
            \textsc{$\mathcal{Q}$-GCN} & 80.34{\scriptsize$\pm$0.07} & 97.87{\scriptsize$\pm$0.01} & 76.33{\scriptsize$\pm$0.01} & 96.15{\scriptsize$\pm$0.01} \\
            \midrule
            \textsc{\modelname{}} & \bf 99.00{\scriptsize$\pm$0.08} & \bf 99.18{\scriptsize$\pm$0.06} & \bf 86.06{\scriptsize$\pm$0.06} & \bf 97.90{\scriptsize$\pm$0.16} \\
            \bottomrule
        \end{tabular}
        }
    \end{minipage}\hfill
    \begin{minipage}[t]{.43\textwidth}
        \centering
        \vspace{-3px}
        \includegraphics[width=\textwidth]{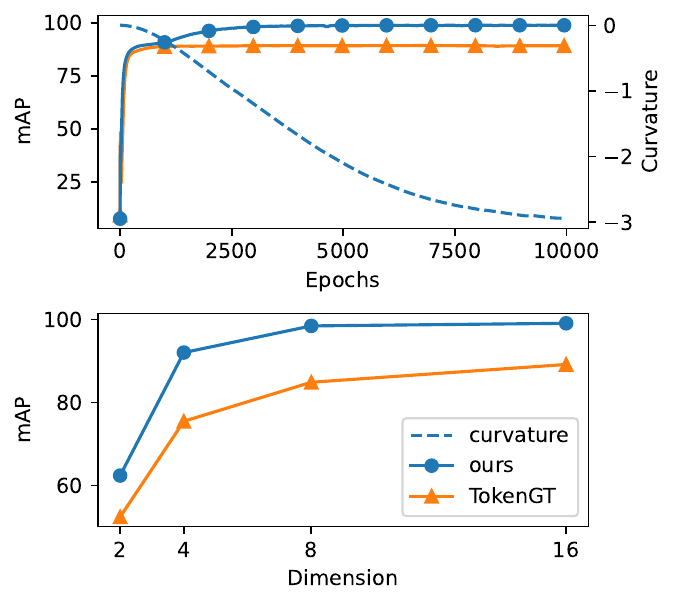}
    \end{minipage}
    \vspace{-5px}
    \caption{\textbf{Left:} Graph reconstruction results. We run each method on 5 different random initializations and report the average mAP score alongside 95\% confidence intervals. \textbf{Upper right:} mAP (solid lines) and curvature (dashed line) of \modelname{} vs. TokenGT during training on Web-Edu. \textbf{Lower right:} Test mAP scores using smaller feature dimensions. Using non-Euclidean geometry leads to better parameter efficiency.}\label{fig:graph_reconstruction}
    \vspace{-15px}
\end{figure}

\paragraph{Training.}
The goal of graph reconstruction is to learn continuous node representations of the given graph that preserve the edge connectivity structure through distances among the learned representations.
Let $\Bh_u$ denote the encoded representation of node $u \in \mathcal{V}$ given a graph $\mathcal{G} = (\mathcal{V}, \mathcal{E})$.
Given the continous representations $\Bh$, we minimize the loss function that aims for preserving the local connections~\cite{(QGCN)xiong2022pseudo}:
\begin{align*}
    \loss_{GR}(\Bh, \mathcal{G}) = \sum_{(u,v) \in \mathcal{E}} \log \dfrac{e^{-d(h_u, h_v)}}{\sum_{v' \in \mathcal{\bar{E}}(u)} e^{-d(h_u,h_{v'})}},
\end{align*}
where $\mathcal{\bar{E}}$ is the set of non-neighbors of node $u$ and $d(\cdot, \cdot)$ is the distance function between the representations on the representation space, the geometry of which depends on the model.
For instance, GCN and HGCN uses Euclidean and hyperbolic space, respectively, while \modelname{} uses the product-stereographic model with curvatures from the last layer. 

For fair comparison, we set the number of layers to one and latent dimension to 16 for all the models.
For $\kappa$-GCN, we use the product of two stereographic models, both of which the curvature is initialized as zero.
For $\mathcal{Q}$-GCN, we test different time dimensions in $\{1, 8, 16\}$, and report the best performance among the models.
For \modelname{}, we use two attention heads with all curvatures initialized as zero.
We train all models for 10k epochs using an Adam optimizer with learning rate $1\mathrm{e}{-2}$. 
The node features are given as one-hot encodings with additional random noise following \cite{(QGCN)xiong2022pseudo}.

\cutsubsectiondown
\paragraph{Results.} 
The table in \autoref{fig:graph_reconstruction} shows the average sectional curvature of each network, and the results in mean average-precision (mAP) which measures the average ratio of nearest points that are actual neighbors of each node.
We find that \modelname{} outperforms the baselines in all datasets.
More importantly, \modelname{} shows significant performance gains compared to Euclidean TokenGT on three networks that are largely hyperbolic.
On Web-Edu with an average sectional curvature of -0.63, \modelname{} shows a 10.5\% gain in mAP against TokenGT, showing that executing attention on the product-stereographic space is especially effective when encoding graphs containing of many non-zero sectional curvatures.



\cutsubsectiondown
\paragraph{Analysis.} For further comparison, we train a single-head \modelname{} and TokenGT on Web-Edu. The upper right plot of \autoref{fig:graph_reconstruction} shows the curvature and mAP scores during training. We find that the curvature is adjusted towards the hyperbolic domain, which matches with the sign of the overall sectional curvature of the Web-Edu network.
The mAP score also converges to a larger mAP as the absolute curvature value deviates further from zero, indicating that the non-Euclidean regime can contain better local optima for graph reconstruction.

Note that non-Euclidean spaces are known to well-embed complex structures in low dimensions, while Euclidean spaces require a large number of dimension to attain reasonable precision~\cite{(hyperbolic_embedding5)pmlr-v80-sala18a}.
Based on this observation, we test whether \modelname{} enjoys better parameter efficiency compared to TokenGT by training two models with varying feature dimensions in $\{2, 4, 8, 16\}$. 
In the lower right plot of \autoref{fig:graph_reconstruction}, we report the performance of TokenGT and \modelname{} post-training.
We observe that \modelname{} preserves the reconstruction performance better as we decrease the dimension from 16, as \modelname{} using only 4 dimensions (92.00 mAP with 12.7k parameters) outperforms TokenGT with $d=16$ (89.13 mAP with 53.6k parameters).


\begin{table}[!t]
    \centering
    \caption{Node classification results. We run each method under 10 different random initializations and report the average F1 scores alongside 95\% confidence intervals.} \label{tab:node_classification} 
    \vspace{.05in}
    \resizebox{\textwidth}{!}{
    \begin{tabular}{l|cccc|cccc}
        \toprule
        Dataset & Texas & Cornell & Wisconsin & Actor & Airport & Citeseer & Pubmed & Cora \\
        $\mathcal{H}(G)$ & 0.11 & 0.13 & 0.20 & 0.22 & 0.72 & 0.74 & 0.80 & 0.81 \\
        \midrule
        \textsc{MLP} & 70.54{\scriptsize$\pm$3.00} & 58.38{\scriptsize$\pm$4.04} & 81.20{\scriptsize$\pm$1.87} & 33.62{\scriptsize$\pm$0.55} & 54.05{\scriptsize$\pm$1.78} & 52.58{\scriptsize$\pm$1.97} & 67.17{\scriptsize$\pm$0.91} & 52.44{\scriptsize$\pm$1.08}\\
        \textsc{GCN} & 57.84{\scriptsize$\pm$1.62} & 47.84{\scriptsize$\pm$1.77} & 45.40{\scriptsize$\pm$2.62} & 27.09{\scriptsize$\pm$0.36} & 92.00{\scriptsize$\pm$0.63} & 71.38{\scriptsize$\pm$0.43} & 78.37{\scriptsize$\pm$0.26} & 80.40{\scriptsize$\pm$0.53} \\
        \textsc{GAT} & 59.46{\scriptsize$\pm$1.12} & 55.14{\scriptsize$\pm$1.80} & 46.20{\scriptsize$\pm$2.30} & 27.43{\scriptsize$\pm$0.23} & 92.35{\scriptsize$\pm$0.36} & 71.70{\scriptsize$\pm$0.28} & 78.14{\scriptsize$\pm$0.31} & 82.29{\scriptsize$\pm$0.46} \\
        \textsc{SAGE} & 68.38{\scriptsize$\pm$3.54} & 70.54{\scriptsize$\pm$2.01} & 78.40{\scriptsize$\pm$0.52} & 36.87{\scriptsize$\pm$0.50} & 93.21{\scriptsize$\pm$0.57} & 70.58{\scriptsize$\pm$0.42} & 77.31{\scriptsize$\pm$0.59} & 78.88{\scriptsize$\pm$0.87} \\
        \textsc{SGC} & 57.57{\scriptsize$\pm$2.96} & 52.97{\scriptsize$\pm$2.87} & 46.40{\scriptsize$\pm$2.01} & 27.14{\scriptsize$\pm$0.46} & 90.48{\scriptsize$\pm$1.01} & \bf 72.11{\scriptsize$\pm$0.38} & 75.11{\scriptsize$\pm$1.27} & 79.68{\scriptsize$\pm$0.65} \\
        \textsc{TokenGT} & 88.65{\scriptsize$\pm$2.06} & 71.62{\scriptsize$\pm$2.13} & 83.00{\scriptsize$\pm$0.65} & 36.59{\scriptsize$\pm$0.89} & 95.90{\scriptsize$\pm$0.59} & 71.23{\scriptsize$\pm$0.51} & \bf 78.93{\scriptsize$\pm$0.27} & 81.42{\scriptsize$\pm$0.79} \\
        \midrule
        \textsc{HGCN} & 54.59{\scriptsize$\pm$3.93} & 55.68{\scriptsize$\pm$1.80} & 55.60{\scriptsize$\pm$2.53} & 28.89{\scriptsize$\pm$0.16} & 92.47{\scriptsize$\pm$0.63} & 69.92{\scriptsize$\pm$0.61} & 75.67{\scriptsize$\pm$0.99} & 80.00{\scriptsize$\pm$0.85} \\
        \textsc{HGNN} & 50.81{\scriptsize$\pm$3.60} & 52.70{\scriptsize$\pm$1.42} & 54.60{\scriptsize$\pm$2.68} & 29.09{\scriptsize$\pm$0.19} & 90.55{\scriptsize$\pm$0.71} & 69.82{\scriptsize$\pm$0.53} & 76.72{\scriptsize$\pm$0.86} & 79.30{\scriptsize$\pm$0.51} \\
        \textsc{HAT} & 82.16{\scriptsize$\pm$2.52} & 70.54{\scriptsize$\pm$1.67} & 81.80{\scriptsize$\pm$1.36} & 38.34{\scriptsize$\pm$0.26} & 92.88{\scriptsize$\pm$0.57} & 68.14{\scriptsize$\pm$0.53} & 77.50{\scriptsize$\pm$0.42} & 79.81{\scriptsize$\pm$0.58} \\
        \midrule
        \textsc{$\kappa$-GCN} & 56.22{\scriptsize$\pm$4.38} & 55.68{\scriptsize$\pm$5.59} & 46.60{\scriptsize$\pm$2.41} & 26.39{\scriptsize$\pm$0.60} & 82.58{\scriptsize$\pm$3.70} & 54.06{\scriptsize$\pm$4.45} & 68.61{\scriptsize$\pm$3.05} & 73.70{\scriptsize$\pm$0.69} \\
        \textsc{$\mathcal{Q}$-GCN} & 51.35{\scriptsize$\pm$3.44} & 55.95{\scriptsize$\pm$2.85} & 52.80{\scriptsize$\pm$2.20} & 28.18{\scriptsize$\pm$0.55} & 91.39{\scriptsize$\pm$1.05} & 66.15{\scriptsize$\pm$0.45} & 77.13{\scriptsize$\pm$0.59} & 79.63{\scriptsize$\pm$0.57} \\
        \midrule
        \textsc{\modelname{}} & \bf 89.19{\scriptsize$\pm$2.37} & \bf 72.16{\scriptsize$\pm$2.96} & \bf 83.60{\scriptsize$\pm$1.14} & \bf 39.61{\scriptsize$\pm$0.54} & \bf 96.01{\scriptsize$\pm$0.55}& 70.03{\scriptsize$\pm$0.71} & 78.52{\scriptsize$\pm$0.58} & \bf 82.32{\scriptsize$\pm$0.70} \\
        \bottomrule
    \end{tabular}
    }
    \vspace{-.15in}
\end{table}

\subsection{Node Classification}
\cutsubsectionup

\paragraph{Datasets.} For node classification we experiment on eight different networks: three WebKB networks (Texas, Cornell, Wisconsin) that connect web-pages via hyperlinks~\cite{(webkb)craven1998learning}, a co-occurrence network from Wikipedia pages related to English films (Actor)~\cite{(actor)tang2009social}, three citation networks (Citeseer, Pubmed, Cora)~\cite{(citation)sen2008collective}, and an airline network (Airport)~\cite{(HGCN)chami2019hyperbolic}. These networks are chosen to test our approach under a wide spectrum of graph homophily $\mathcal{H}(G)$, which measures the ratio of edges that connect nodes that share the same label~\cite{(edge_homophily)zhu2020beyond}. In other words, a hetereophilic graph with small graph homophily requires capturing long-range interactions for proper labeling, which is naturally difficult for message passing-based approaches with small receptive fields. More detailed statistics on the networks can be found in \autoref{apx:dataset_statistics}.

\cutsubsectiondown
\paragraph{Training.} For all methods, we fix the embedding dimension to 16 and train each model to minimize the cross-entropy loss using an Adam optimizer with a learning rate of $1\mathrm{e}{-2}$. For models that use learnable curvatures (\textit{i.e.}, HGCN, $\kappa$-GCN and \modelname{}), we use a learning rate of $1\mathrm{e}{-4}$ for the curvatures. The optimal number of layers, activation function, dropout rate, and weight decay of each method are chosen via grid search on each dataset. Details on the hyperparameter search-space and dataset splits can be found in Appendix \ref{apx:node_classification_details}.

\cutsubsectiondown
\paragraph{Results.} \autoref{tab:node_classification} shows the results from node classification. Overall, our method shows best accuracy on 6 out of 8 datasets, showing that \modelname{} is effective across networks with various graph homophily. In case of hetereophilic networks, we find that the small receptive fields of message-passing GCNs are extremely inadequate, often being outperformed by MLPs that completely ignore the graph connectivity. On the other hand, \modelname{} consistently outperforms MLP as well as GCNs, due to being able to exchange information through long distances via global-attention. It also significantly outperforms TokenGT by 8.3\% on Actor, showing that adjusting the geometry towards non-Euclidean can further enhance predictive performance. In homophilic networks where message-passing is more well-suited, \modelname{} shows competitive performance against GCN baselines. This is expected as \modelname{} enjoys the same capacity as TokenGT to mimic any order-2 equivariant bases~\cite{(TokenGT)kim2022pure}, which includes local message-passing, through attention score computation.




\cutsectionup
\section{Conclusion}
\cutsectiondown

We propose \modelname{}, a natural generalization of the Transformer architecture towards the non-Euclidean domain. When combined with the graph tokenization technique of TokenGT~\cite{(TokenGT)kim2022pure}, our model can embed graphs with less distortion and higher parameter-efficiency than its Euclidean counterpart by operating on the product-stereographic model with learnable curvatures. We also show that our model outperforms existing hyperbolic and mixed-curvature message-passing GCN baselines on node classification via global-attention that can capture long-range interactions. By linearizing the cost of self-attention through kernelized approximation, \modelname{} runs in cost linear to the number of nodes and edges, allowing practical use on large-scale networks. 
For future work, we plan to extend towards heterogeneous manifolds~\cite{(heterogeneous)giovanni2022heterogeneous} with input-dependent sectional curvatures as well as optimize Stereographic operations towards better stability and efficiency under machine precision. As we propose a foundational generalization of the Transformer framework, we do not expect any immediate negative societal impact from this work. 



\bibliographystyle{plainnat}
\bibliography{neurips_2023.bib}

\newpage
\appendix
\newcounter{questioncntr}
\newcommand*{\question}{%
    \stepcounter{questioncntr}%
    \textbf{Q\arabic{questioncntr}: }%
}
\newcommand*{\answer}{%
    \textbf{A\arabic{questioncntr}: }%
}

\setcounter{section}{16}

\section*{Answers to Potential Questions}
\cutsectionup

To provide a better understanding of our draft, we start the supplementary material by providing answers to potential questions on the overall motivation of our work, our proposed methodology, and presented empirical results. We hope most questions during review can be answered in this section, and would be happy to clarify any further questions during the author response period as well. Further supplementary material can be found in the following sections.


\question \textbf{Why should we consider mixed-curvature spaces for graph representation learning?}\\
\answer Previous works have shown that graphs with both hyperbolic (\textit{e.g.} hierarchical trees) and spherical (\textit{e.g.} cycles) structures require both curvature spaces to be embedded accurately~\cite{(mixed_curvature)gu2018learning, (kappa-GCN)bachmann2020constant}. Based on our sectional curvature estimations in \autoref{fig:sectional_curvatures}, we find that this is often the case in real-world networks, showing sectional curvatures with both negative and positive signs. As the stereographic model can universally model both spherical and hyperbolic spaces, we consider it to be a good fit for learning such networks. The benefit of mixed-curvature spaces is also evident in our results on graph reconstruction, as \modelname{} outperforms hyperbolic baselines (\textit{i.e.} HGCN~\cite{(HGCN)chami2019hyperbolic}, HGNN~\cite{(HGNN)liu2019hyperbolic}, HAT~\cite{(HAT)zhang2021hyperbolic}).

\question \textbf{What advantages does the product-steregraphic model have over the pseudo-Riemannian manifold used in $\mathcal{Q}$-GCN~\cite{(QGCN)xiong2022pseudo}?}\\
\answer While the pseudo-Riemannian manifold with an indefinite metric can model both spherical and hyperbolic spaces by containing both as submanifolds, the $\mathcal{Q}$-GCN architecture requires setting the time-dimension of the manifold as hyperparameter~\cite{(QGCN)xiong2022pseudo} (a pseudo-hyperboloid with larger time-dimension makes it more similar to a spherical manifold). According to the results in \cite{(QGCN)xiong2022pseudo}, we find that the downstream performance is sensitive to the time-dimension, which implies extensive hyperparameter tuning for quality predictions. To make things worse, the number of possible time-dimensions increases linearly to the embedding dimension of choice, which can make the tuning process intractable when scaling up towards large dimensions. 

On the other hand, the product-stereographic model used in \modelname{} does not require any such hyperparameter tuning. While the initial curvatures may be of concern during model initialization, we find that \modelname{} performs well under initially setting all curvatures to zero, thereby starting from a Euclidean space and gradually tuning the curvatures to fit the input graph.


\question \textbf{How is \modelname{} different from Hyperbolic Attention Network~\cite{(hyp-attn)gulcehre2018hyperbolic}?}\\
\answer Hyperbolic Attention Network (HAtt)~\cite{(hyp-attn)gulcehre2018hyperbolic} performs global-attention by computing the attention scores based on the hyperbolic distances between query- and key-vectors. Our \modelname{} instead extends the dot-product attention from vanilla Transformer~\cite{(Transformer)vaswani2017attention}, thereby inheriting the same theoretical expressiveness on universal approximability~\cite{(universal_approximation)yun2019transformers} through a natural generalization to mixed-curvature spaces. Furthermore, our method can tune the curvatures based on the data and task and can represent both positive and negative curvatures, whereas HAtt uses the hyperboloid and the Klein model and thus its representation space is limited within the hyperbolic domain.

\question \textbf{How is the running cost \modelname{} compared vs. TokenGT~\cite{(TokenGT)kim2022pure} and other baselines~\cite{(QGCN)xiong2022pseudo, (kappa-GCN)bachmann2020constant}?}\\
\answer For comparing the computational cost of \modelname{} against our baselines, we measure the runtime and peak memory use during inference on the four networks used in our graph reconstruction datasets. Note the network statistics are presented in \autoref{tab:graph_reconstruction_dataset_statistics}. 

\autoref{fig:computational_cost} shows the computational cost measurements.
Compared to TokenGT, \modelname{} essentially does not use more memory. 
Furthermore, both time and memory cost of \modelname{} are far below those of $\mathcal{Q}$-GCN, which demonstrates better utility of closed-form operations on the $\kappa$-stereographic model compared to those on the pseudo-Riemannian manifold.

\begin{figure}[!ht]
    \begin{minipage}[t]{.49\textwidth}
        \centering
        \vspace{0pt}
        \resizebox{\textwidth}{!}{\begin{tabular}{l|cccc}
        \toprule
        Dataset & Web-Edu & Power & Facebook & Bio-Worm \\
        \midrule
        GCN & 0.71 & 2.54 & 2.71 & 1.45\\
        GAT & 7.90 & 7.35 & 11.24 & 7.86\\
        SAGE & 10.90 & 9.13 & 22.20 & 7.66\\
        SGC & 1.11 & 2.35 & 3.12 & 2.02\\
        TokenGT & 10.10 & 8.93 & 57.09 & 50.28\\
        \midrule
        HGCN & 8.55 & 10.64 & 9.50 & 9.06\\
        HGNN & 5.90 & 6.51 & 9.57 & 9.59\\
        HAT & 17.50 & 20.11 & 20.11 & 16.15\\
        \midrule
        $\kappa$-GCN & 7.91 & 10.68 & 8.84 & 8.03\\
        $\mathcal{Q}$-GCN & 72.09 & 75.36 & 74.62 & 72.42\\
        \midrule
        \modelname{} & 22.72 & 21.64 & 70.94 & 66.44\\
        \bottomrule
        \end{tabular}
        }
    \end{minipage}\hfill
    \begin{minipage}[t]{.49\textwidth}
        \centering
        \vspace{0pt}
        \resizebox{\textwidth}{!}{\begin{tabular}{l|cccc}
        \toprule
        Dataset & Web-Edu & Power & Facebook & Bio-Worm \\
        \midrule
        GCN & 37.72 & 96.32 & 74.63 & 30.26\\
        GAT & 37.99 & 96.49 & 80.46 & 35.52\\
        SAGE & 145.05 & 377.39 & 257.01 & 86.96\\
        SGC & 37.89 & 96.60 & 74.86 & 30.39\\
        TokenGT & 73.64 & 190.38 & 437.80 & 352.60 \\
        \midrule
        HGCN & 144.51 & 376.67 & 252.64 & 83.18\\
        HGNN & 144.51 & 376.67 & 252.64 & 83.18\\
        HAT & 181.01 & 413.18 & 289.14 & 119.69\\
        \midrule
        $\kappa$-GCN & 74.80 & 133.70 & 113.11 & 68.32\\
        $\mathcal{Q}$-GCN & 361.03 & 883.21 & 600.42 & 219.70\\
        \midrule
        \modelname{} & 73.64 & 190.38 & 444.97 & 358.60\\
        \bottomrule
        \end{tabular}
        }
    \end{minipage}
    \caption{Average runtime (left, in ms) and peak memory (right, in MB) estimation during inference. Each table shows average results over 20 different runs.}\label{fig:computational_cost}
    \vspace{-20px}
\end{figure}



\question \textbf{What are the features used in the graph reconstruction and node classification experiments?}\\
\answer For graph reconstruction, we use a one-hot encoding as node representations, and thus each node embedding is learned independently. Specifically for TokenGT and \modelname{} where the sequence includes edge tokens as well, we do not provide any edge features, and thus the edge tokens only contain the positional encoding of the graph and its type identifier (more details can be found in \autoref{apx:architecture}).

For node classification, we use the node features provided from each dataset. The WebKB networks (Cornell, Texas, Wisconsin) use bag-of-words representations of each web-page as the input node features. Actor uses a set of keywords from the Wikipedia page pertaining to each actor-node. Airport uses node features that contain geographic location of each airport as well as GDP of the country in which the airport is located, following \cite{(HGCN)chami2019hyperbolic}. Citation networks (Citeseer, Cora, Pubmed) use bag-of-words representations of each paper. Same as in graph reconstruction, we do not endow additional link features for TokenGT and \modelname{} other than the positional and token-type information.

\question \textbf{Baselines such as GIL~\cite{(GIL)zhu2020graph} and FMGNN~\cite{(FMGNN)deng2023fmgnn} are missing in the experiments.}\\
\answer We have tested GIL in our experiments, but found that the published code shows inconsistent performance across random seeds when compared with the original paper~\cite{(GIL)zhu2020graph}. We were not able to reproduce results at the time of writing, and hence leave the comparison vs. GIL as future work. For FMGNN~\cite{(FMGNN)deng2023fmgnn}, we have not yet found published official code possibly due to the work being published recently, and thus were not able to include the method as our baseline. Nonetheless, we believe that these methods are still limited by their small receptive fields as they lie within the message-passing paradigm, and expect \modelname{} to outperform on heterophilic graphs as in our node classification experiments.

\question \textbf{$\kappa$-GCN~\cite{(kappa-GCN)bachmann2020constant} and $\mathcal{Q}$-GCN~\cite{(QGCN)xiong2022pseudo} are outperformed by Euclidean baselines in graph reconstruction. Why is this so?}\\
\answer We found that while the two methods are able to leverage non-Euclidean geometry, their performances are sensitive to different architectural choices. For instance, we found that $\kappa$-GCN with all curvatures initialized at zero does not perform well on real-world graphs, potentially due to issues in optimization. Similarly, $\mathcal{Q}$-GCN performs poorly when using a larger embedding dimension of 16 rather than 10 (as per their original paper), and is also very sensitive to the time-dimension parameter. On the other hand, \modelname{} performs well under a simple initialization of zero curvature without any additional hyperparameter. While we could perform extensive tuning on the curvature initializations and time-dimensions for $\kappa$-GCN and $\mathcal{Q}$-GCN, respectively, we consider this to be outside the scope of our paper.


\setcounter{section}{0}

\section{Riemannian Operations on the Stereographic Model}\label{apx:stereographic}
In this section, we introduce closed-form equations of Riemannian operations on the stereographic model.
We first define the $\tan_\kappa$ and $\sin_\kappa$ as:
\begin{align*}
    \tan_\kappa(x) = \begin{cases}
        \frac{1}{\sqrt{\kappa}}\tan(\sqrt{\kappa}x), & \kappa > 0 \\
        x, & \kappa = 0 \\
        \frac{1}{\sqrt{-\kappa}}\tanh(\sqrt{-\kappa}x), & \kappa < 0.
    \end{cases}
    ,\quad\quad &
    \sin_\kappa(x) = \begin{cases}
        \frac{1}{\sqrt{\kappa}}\sin(\sqrt{\kappa}x), & \kappa > 0 \\
        x, & \kappa = 0 \\
        \frac{1}{\sqrt{-\kappa}}\sinh(\sqrt{-\kappa}x), & \kappa < 0.
    \end{cases}
\end{align*}

Based on the mobius addition and $\tan_\kappa$, we can define the Riemannian operations of the stereographic model as shown in \autoref{tab:stereographic_ops}.

\begin{table}[!h]
    \centering
    \caption{
    Closed-forms of the Riemannian operations of the stereographic model. As the curvature $\kappa$ converges to zero, the Riemannian operations recover the Euclidean operations.
    } 
    \label{tab:stereographic_ops} 
    \vspace{3px}
    \resizebox{.9\textwidth}{!}{
    \begin{tabular}{l l c}
        \toprule
        Operations & \multicolumn{1}{c}{$\kappa \in \R$} & $\kappa \rightarrow 0$ \\
        \midrule
        Distance & $d_\kappa(\Bx, \By) = 2 \tan_{\kappa}^{-1}(\Vert -\Bx \oplus_{\kappa} \By \Vert)$ & $2\Vert \By - \Bx \Vert$ \\
        Exponential map & $\exp^{\kappa}_{\Bx}(\Bv) = \Bx \oplus_{\kappa} \left(\tan_\kappa\left(\frac{\sqrt{\vert \kappa \vert}\lambda_{\Bx}^\kappa \Vert \Bv \Vert}{2} \right) \frac{\Bv}{\Vert \Bv \Vert} \right)$ & $\Bx + \Bv$\\
        Log map & $\log^\kappa_{\Bx}(\By) = \frac{2}{\sqrt{\vert \kappa \vert} \lambda_{\Bx}^\kappa} \tan_\kappa^{-1}(\Vert -\Bx \oplus_\kappa \By \Vert) \frac{-\Bx \oplus_\kappa \By}{\Vert -\Bx \oplus_\kappa \By \Vert}$ & $\By - \Bx$\\
        Parallel transport & $\mathrm{PT}_{\Bx \rightarrow \By}^\kappa (\Bv) = (-(\By \oplus_\kappa -\Bx) \oplus_\kappa (\By \oplus_\kappa (-\Bx \oplus_\kappa \Bv))) \cdot \frac{\lambda_{\Bx}^\kappa}{\lambda_{\By}^\kappa}$ & $\Bv$ \\
        \bottomrule
    \end{tabular}
    }
    \vspace{-10px}
\end{table}

\section{Details of \modelname{}}\label{apx:architecture}

\subsection{Tokenization Procedure of TokenGT}
\cutsubsectionup
In order to learn graph-structured data with \modelname{}, we borrow the tokenization technique proposed by TokenGT~\cite{(TokenGT)kim2022pure}. Let graph $\mathcal{G} = (\mathcal{V}, \mathcal{E})$ be an input graph with $N$ nodes in node-set $\mathcal{V}$, $M$ edges in edge-set $\mathcal{E}$, and respective features $\BX^\mathcal{V} \in \R^{N \times d}$, $\BX^\mathcal{E} \in \R^{M \times d}$. Then, we tokenize the graph into a sequence by treating each node and edge as an independent token, and augment each token embedding with 1) node identifiers $\BP \in \R^{N \times d}$ that serve as positional encoding and 2) type identifiers $\BE \in \R^{2 \times d}$ that allows the model to distinguish between node- and edge-tokens.
\begin{align*}
    \BX_u &\mapsto \BX_u + \BP_u + \BP_u + \BE_0 \text{ for each node } u \in \mathcal{V}\\
    \BX_{(u,v)} &\mapsto \BX_{(u,v)} + \BP_u + \BP_v + \BE_1 \text{ for each edge } (u,v) \in \mathcal{E}.
\end{align*}
The node identifiers in $\BP$ are obtained from top-$k$ eigenvectors of the graph Laplacian $\BI - \BD^{-1/2} \BA \BD^{-1/2}$ where $\BA$ and $\BD$ denote the adjacency and degree matrices, respectively. The type identifiers $\BE$ are set as trainable parameters. 

TokenGT feeds this sequence into a pure Euclidean Transformer, an approach proven to pass the 2-dimensional Weisfeiler-Lehman (2-WL) graph isomorphism test and surpass the theoretical expressivity of message-passing GCNs~\cite{(TokenGT)kim2022pure, (expressivity)maron2019provably}. In our work, we encode the input sequence through \modelname{} instead, such that nodes and edges exchange information globally on the product-stereographic space.

\subsection{Overall Architecture}

\begin{figure}[!h]
    \centering
    \includegraphics[trim={0 2cm 2.3cm 1cm},clip,width=.95\textwidth]{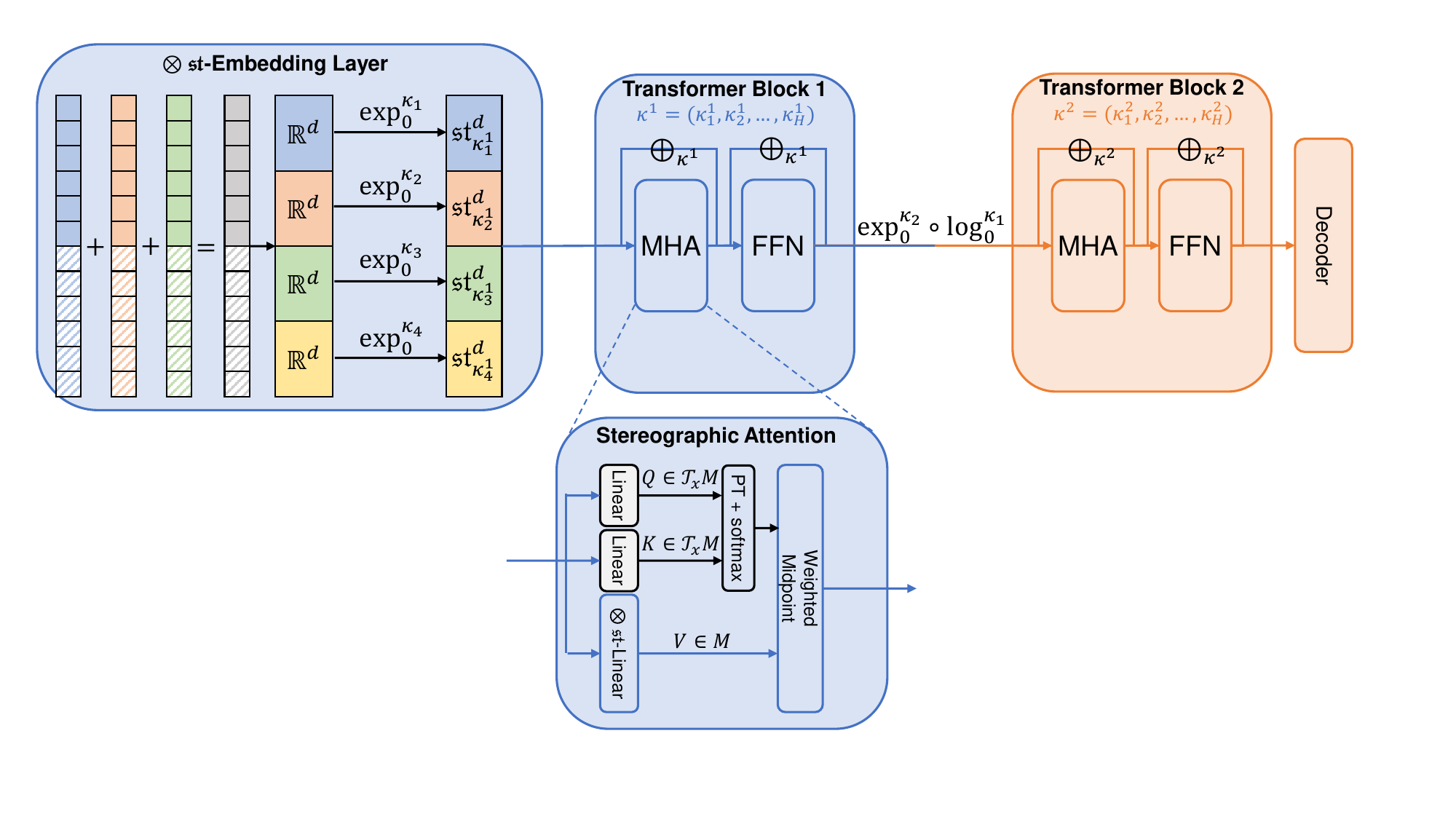}
    \caption{Illustration of the overall \modelname{} architecture. Each layer of \modelname{} is endowed with a product-streographic space with a curvature assigned to each attention head. The embedding layer applies exp-mapping to the token embeddings using the curvatures of the first layer. The decoder uses the curvatures of the last layer. In-between, we assume the tangent space at the origin is shared between the two product-stereographic spaces, and translate the embeddings via $\exp_{\Bzero}^{\otimes\Bkappa_{l+1}} \circ \log_{\Bzero}^{\otimes\Bkappa_l}$.}
    \label{fig:overall_architecture}
\end{figure}

For further guidance, we provide a more detailed illustration of our \modelname{} architecture in \autoref{fig:overall_architecture}.

\newpage
\section{Derivation of Linearized Stereographic Attention}\label{apx:linearized}

The equation below shows the derivation of our steregraphic attention with cost linear to the sequence length. The key step is the second step where the Euclidean inner product at the tangent space of the origin is approximated as a dot product of kernel-mappings, as used in \cite{(linearized_attention)katharopoulos2020transformers}.

\begin{align*}
    \textrm{LinearizedAggregate}_{\kappa}(\BV, \Balpha)_i &= 
    \dfrac{1}{2} \otimes_{\kappa} \left(\sum_{j=1}^n \dfrac{\langle \tilde{\BQ}_i, \tilde{\BK}_j \rangle_{\Bzero} \lambda_{\BV_j}^\kappa}{\sum_{k=1}^n \langle \tilde{\BQ}_i, \tilde{\BK}_k \rangle_{\Bzero} (\lambda_{\BV_k}^\kappa - 1)}\BV_j \right) \\
    &\approx \dfrac{1}{2} \otimes_{\kappa} \left(\sum_{j=1}^n \dfrac{ \phi(\tilde{\BQ}_i)\phi(\tilde{\BK}_j)^T (\lambda^\kappa_{\BV_j} - 1)}{\sum_{k=1}^n \phi(\tilde{\BQ}_i)\phi(\tilde{\BK}_k)^T (\lambda_{\BV_k}^\kappa - 1)} \cdot \dfrac{\lambda^\kappa_{\BV_j}}{\lambda_{\BV_j} - 1} \BV_j \right) \\
    &= \dfrac{1}{2} \otimes_{\kappa} \left[\phi(\tilde{\BQ})\left(\phi'(\tilde{\BK})^T \tilde{\BV}\right)\right]_i
\end{align*}

\section{Dataset Statistics}\label{apx:dataset_statistics}

Here we provide basic statistics on each dataset used in our experiments.

\begin{table}[!ht]
    \centering
    \caption{Dataset statistics for graph reconstruction.} \label{tab:graph_reconstruction_dataset_statistics} 
    \resizebox{.5\textwidth}{!}{\begin{tabular}{l|cccc}
        \toprule
        Dataset & Web-Edu & Power & Facebook & Bio-Worm \\
        \midrule
        \# nodes & 3,031 & 4,941 & 4,039 & 2,274 \\
        \# edges & 6,474 & 6,594 & 78,328 & 88,234 \\
        \bottomrule
    \end{tabular}
    }
\end{table}

\begin{table}[!ht]
    \centering
    \caption{Dataset statistics for node classification.} \label{tab:transductive_dataset_statistics} 
    \vspace{.1in}
    \resizebox{.8\textwidth}{!}{\begin{tabular}{l|cccc|cccc}
        \toprule
         & \multicolumn{4}{c|}{\bf Heterophilic} & \multicolumn{4}{c}{\bf Homophilic}\\
        Dataset & Texas & Cornell & Wisconsin & Actor & Airport & Citeseer & Pubmed & Cora \\
        \midrule
        $\mathcal{H}(G)$ & 0.11 & 0.13 & 0.20 & 0.22 & 0.72 & 0.74 & 0.80 & 0.81\\
        \# nodes & 183 & 183 & 251 & 7,600 & 3,188 & 3,327 & 19,717 & 2,708 \\
        \# edges & 325 & 298 & 515 & 30,019 & 18,631 & 4,732 & 44,338 & 5,429 \\
        \# features & 1,703 & 1,703 & 1,703 & 932 & 4 & 3,703 & 500 & 1,433 \\
        \# classes & 5 & 5 & 5 & 5 & 4 & 6 & 3 & 7\\
        \bottomrule
    \end{tabular}
    }
\end{table}

\section{Experimental details}\label{apx:experimental_details}

\subsection{Graph Reconstruction}\label{apx:graph_reconstruction_details}

For graph reconstruction, we use a single-layer architecture with 16 embedding dimensions for all variants without hyperparameter tuning for fair comparison. We feed the entire network at each training step with no weight decay or dropout, as the objective is to simply embed nodes on the representation space such that the distances on the feature space are well-aligned to the actual graph topology. For the smaller dimension experiment shown in the upper right of \autoref{fig:graph_reconstruction}, we find that removing layer normalization helps performance, and hence we remove for both TokenGT and \modelname{}.

\subsection{Node Classification}\label{apx:node_classification_details}

\paragraph{Hyperparameter search space.}

We share the hyperparameter search space used for node classification in \autoref{tab:hyperparameters}. The lower two parameters are exclusive for only TokenGT and \modelname{}. Note that the number of hops denotes the number of message-passing steps used to mix input node features. This is based on the previous observation that TokenGT benefits manually injecting a sparse equivariant basis to mix node features, due to TokenGT not being able to satisfy the orthogonality constraint when the number of Laplacian eigenvectors used for positional encoding is significantly less than the number of nodes in the graph~\cite{(TokenGT)kim2022pure}.

\begin{table}[!h]
    \centering
    \vspace{-10px}
    \caption{Hyperparameter search space used for node classification.} \label{tab:hyperparameters} 
    \vspace{.05in}
    \resizebox{.7\textwidth}{!}{
    \begin{tabular}{lc}
        \toprule
        Parameter & Search Space \\
        \midrule
        Number of layers & \{1, 2, 3\}\\ 
        Number of heads & \{1, 2, 4\}\\
        Weight decay & \{0, 0.0001, 0.0005, 0.001\} \\
        Dropout & \{0, 0.1, 0.2, 0.3, 0.4, 0.5, 0.6, 0.7\}\\
        Activation & \{ReLU, ELU, Tanh, Sigmoid\}\\
        \midrule
        Number of hops & \{0, 10, 20, 30\}\\
        Number of Laplacian eigenvectors & \{4, 8, 16, 32\}\\
        \bottomrule
    \end{tabular}
    }
    \vspace{-.15in}
\end{table}

\paragraph{Dataset splits.} We use the same data splits used in previous work~\cite{(HGCN)chami2019hyperbolic, (QGCN)xiong2022pseudo}. 
The training set for the node classification task of the 3 citation networks is consisted of 20 nodes per class and the validation/test set is created by sampling 500/1000 nodes from the remaining nodes.
For the Airport network, we use a 70/15/15 split for training, validating, and testing, respectively. 
For the remaining four heterophilic networks, we use a 60/20/20 split. 

\section{Graph Sectional Curvature}\label{apx:sectional_curvature}

\subsection{Computation}
Given an unweighted graph $G(V, E)$, we can compute the graph sectional curvature of the node $m$ and its two neighbors $b, c$ as:
\begin{equation}\label{eqn:sectional_curvature}
    K_G(m; b, c) = \frac{1}{\vert V \vert} \sum_{a \in V} d_G(a, m)^2 + \frac{d_G(b, c)^2}{4} - \frac{d_G(a, b)^2 + d_G(a, c)^2}{2},
\end{equation}
where $d_G(x, y)$ is the shortest path on the graph between node $x$ and $y$.
The graph sectional curvature is known to indicate specific structures such as line, trees, and cycles~\cite{(mixed_curvature)gu2018learning}.

\subsection{Sectional Curvature of the Real-World Graphs}

\begin{figure}[!h]
    \centering
    \includegraphics[width=\textwidth]{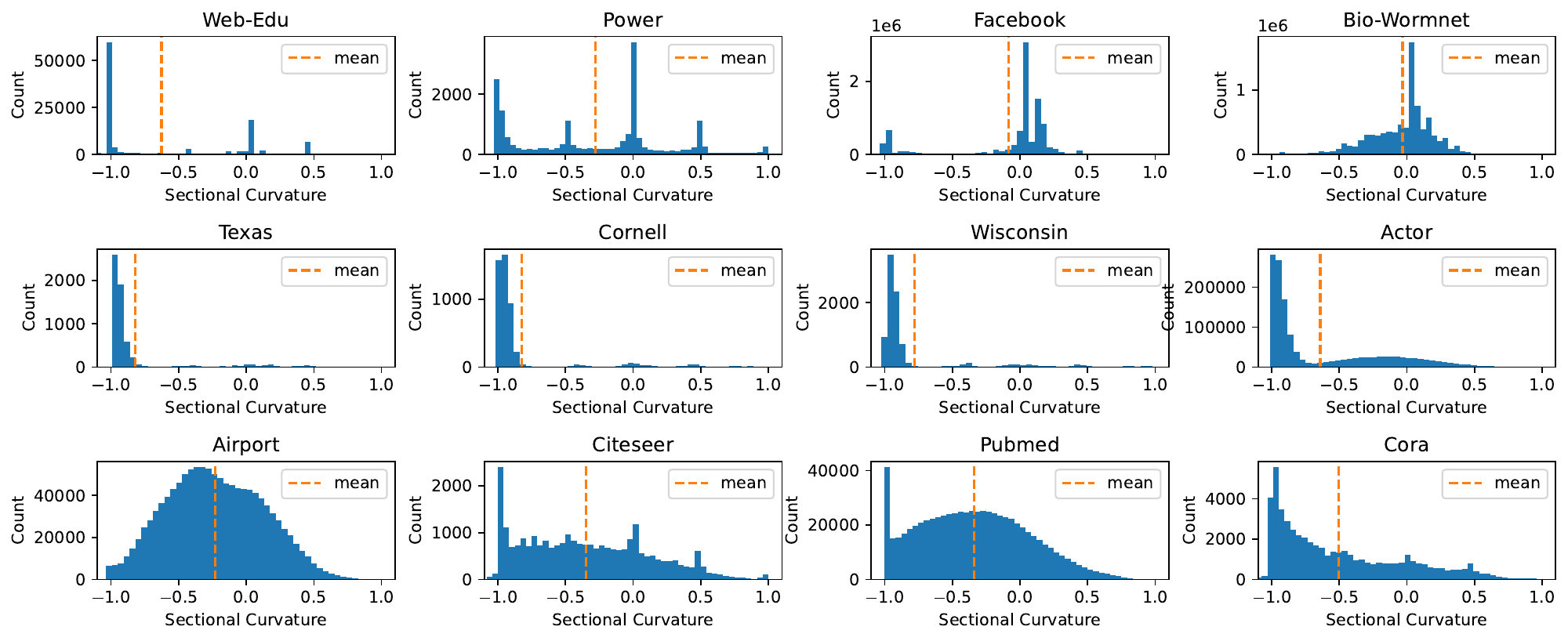}
    \caption{Sectional curvature histograms of all networks used in our experiments.}
    \label{fig:sectional_curvatures}
\end{figure}

Figure~\ref{fig:sectional_curvatures} shows the histograms of sectional curvatures computed by Equation~\ref{eqn:sectional_curvature} in every real-world graph we use in the paper.

\newpage
\section{Visualization}

\begin{figure}[ht]
    \begin{minipage}[t]{.5\textwidth}
        \centering
        \includegraphics[trim={1cm 1cm 1cm 1cm},clip,width=\textwidth]{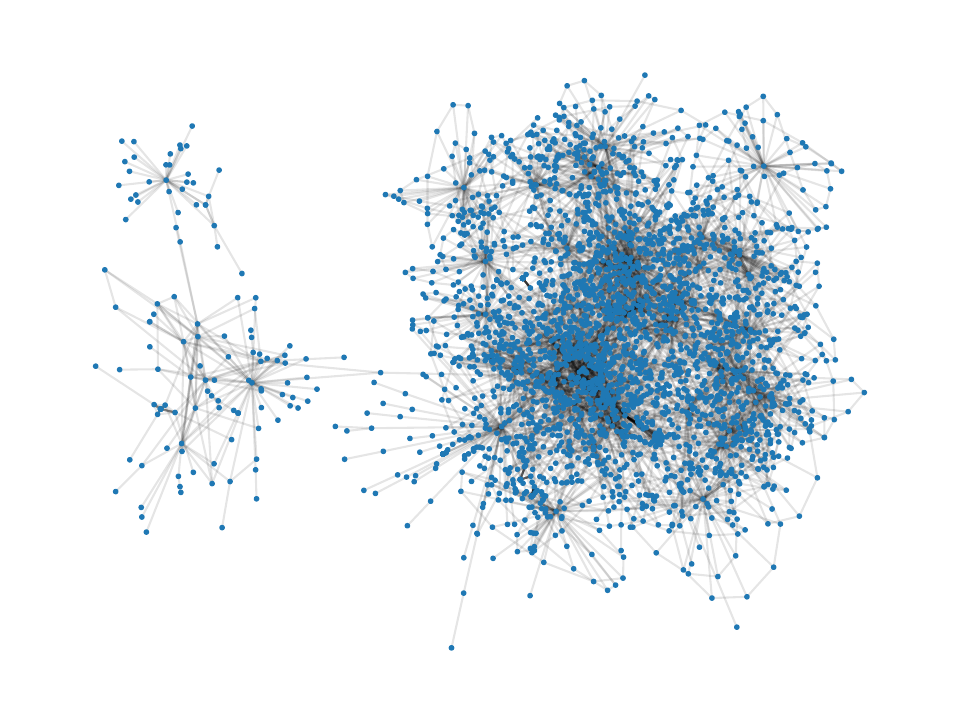}
    \end{minipage}\hfill
    \begin{minipage}[t]{.5\textwidth}
        \centering
        \includegraphics[trim={1cm 1cm 1cm 1cm},clip,width=\textwidth]{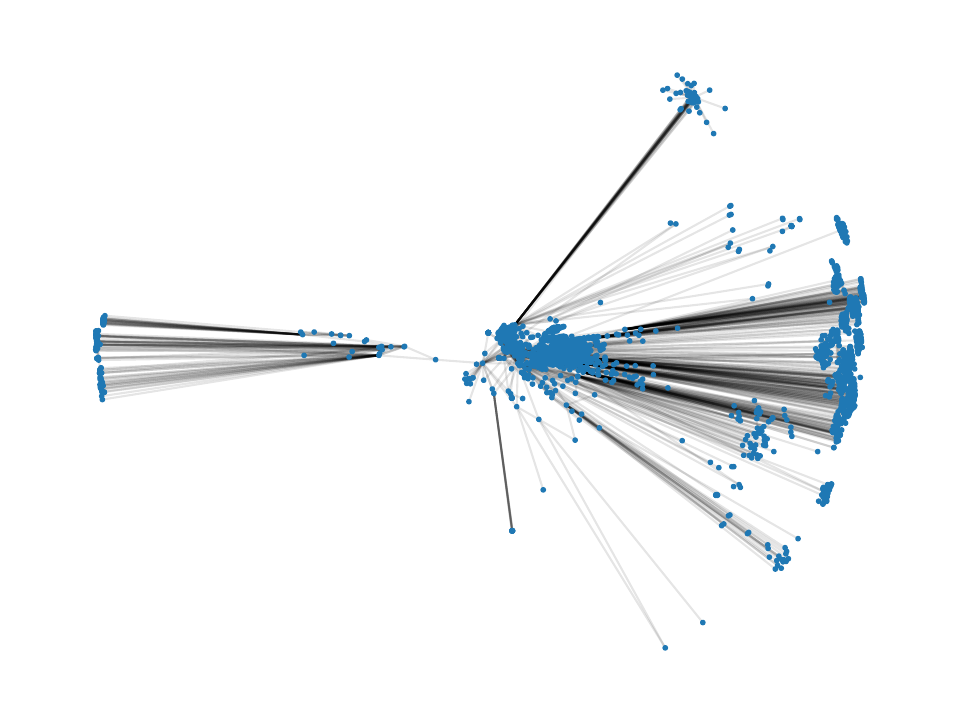}
    \end{minipage}
    \caption{PCA visualizations of Web-Edu node embeddings learned by TokenGT (left) and \modelname{} (right). Restricting the model to the Euclidean space $\mathbb{E}$ leads to convoluted structures and hence suboptimal graph distortion. \modelname{}, on the other hand, adjusts itself towards the hyperbolic product-stereographic space $\mathfrak{st}_{-0.5}^{8} \times \mathfrak{st}_{-3.9}^{8}$, and captures the hierarchical structure of the graph more accurately.}\label{fig:graph_reconstruction_vis}
\end{figure}

\autoref{fig:graph_reconstruction_vis} shows example visualizations of embeddings trained via graph reconstruction on the Web-Edu network. We visualize embeddings from TokenGT and \modelname{} by running PCA on the embeddings directly (for TokenGT), or log-mapping the embeddings to the weighted midpoint first, then applying PCA to reduce the dimension (for \modelname{}).

\end{document}